\documentclass[10pt, a4paper]{article}
\usepackage{lrec}
\usepackage{multibib}
\newcites{languageresource}{Language Resources}
\usepackage{graphicx}
\usepackage{tabularx}
\usepackage{soul}

\usepackage{epstopdf}
\usepackage[latin1]{inputenc}

\usepackage{hyperref}
\usepackage{xstring}
\usepackage{verbatim}

\usepackage[font=small,labelfont=bf,tableposition=top]{caption}
\usepackage{blindtext}
\usepackage{cuted}

\title{Text Annotation Graphs: Annotating Complex Natural Language Phenomena}

\name{Angus G. Forbes$^{\ast}$, Kristine Lee$^{\dagger}$, Gus Hahn-Powell$^{\ddagger}$, \\ {\bf \large Marco A. Valenzuela-Esc\'{a}rcega$^{\ddagger}$, Mihai Surdeanu$^{\ddagger}$ }\vspace{0.25cm} }

\address{$^{\ast}$University of California, Santa Cruz, angus@ucsc.edu\\ $^{\dagger}$University of Illinois at Chicago, khlee2@uic.edu \\$^{\ddagger}$University of Arizona, \{hahnpowell,marcov,msurdeanu\}@email.arizona.edu\\}

\abstract{
This paper introduces a new web-based software tool for annotating text, \textit{Text Annotation Graphs}, or \textit{TAG}. It provides functionality for representing complex relationships between words and word phrases that are not available in other software tools, including the ability to define and visualize relationships between the relationships themselves (semantic hypergraphs). Additionally, we include a visualization mode in which annotation subgraphs, or semantic summaries, are used to show relationships outside of the sequential context of the text itself. These subgraphs can be used to quickly find similar structures within the current document or external annotated documents. \textit{TAG} was initially developed to support information extraction tasks on a large database of biomedical articles. However, our software is flexible enough to support a wide range of annotation tasks for many domains. Examples are provided that showcase \textit{TAG}'s capabilities on morphological parsing and event extraction tasks. 
\\ \newline  }

\begin{document}

\maketitleabstract


   \begin{strip}
 \vspace{-1.25cm}
 
     \centering\noindent
     \includegraphics[width=6.6in]{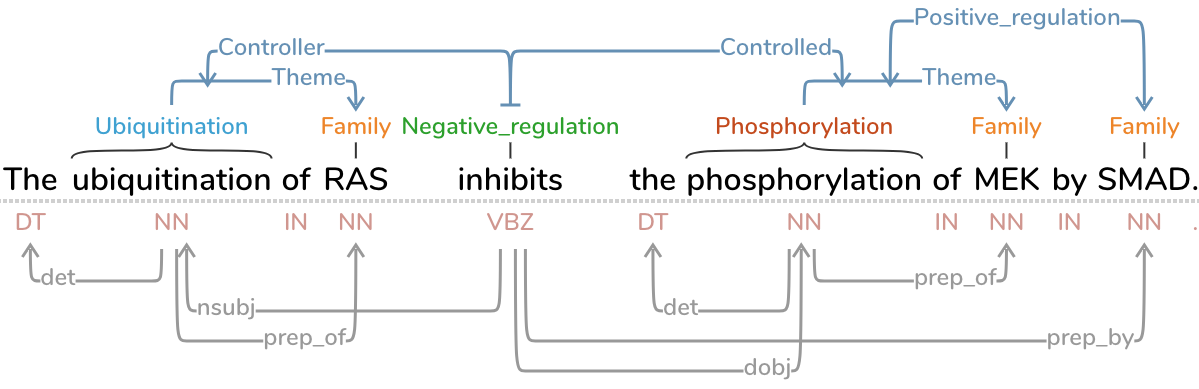} 
     \captionof{figure}{A screen capture of the annotation panel in the \textit{TAG} software. Here a single sentence of biomedical text is shown, with both semantic (top) and syntactic (bottom) relationships toggled on.}
     \label{onerow}
   \end{strip}


\section{Introduction}

According to systems biologist Arthur Lander, the dominant icon of our current era is the ``hairball''~\cite{Lander2010}, a visual depiction of a network of such complexity that the edges overlap so densely as to obscure the meaning of the very relationships it is meant to explain. Part of the reason for this shift from a more straightforward symbol (such as the double helix, which, while still complex, indicates a more orderly, and potentially predictable machinery) to the hairball, Lander writes, is the rise of the computational tools that make it possible to generate vast amounts of ``big data,'' as well as the realization that thorny problems generally involve the intersections between data drawn from multiple, interacting systems. 

\vspace{.1cm}

Contemporary information extraction approaches seek to capture complex natural language phenomena, such as those found in biomedical literature~\cite{bjorne2010complex,Valenzuela2015domain}. These biomedical articles present new knowledge regarding biological processes which can be usefully described via a semantic graph that represents the event structure of biochemical reactions. Moreover, these approaches can be used to detect predicates (triggers), relations, and events within individual articles in order to find connections to other corpora and to draw inferences about relationships between texts. 
\vspace{.1cm}

We introduce a novel web-based annotation software application, \textit{Text Annotation Graphs}, or \textit{TAG}, that provides a rich set of tools for viewing and editing annotations that capture complex natural language phenomena, such as deeply nested event structures (Figs.~\ref{onerow} and~\ref{brat_vs_tag_ea}), syntactic dependencies (bottom portion of Fig.~\ref{onerow}), undirected relations, coreference resolution, and morphological parses (Fig.~\ref{morphTrees}). \textit{TAG} provides easily readable output for complex events and can be used to illustrate non-projective representations and even link predicate-less relations. It supports multiple languages and data formats, allows users to view layers of annotations simultaneously, and includes interactive features for filtering and editing annotations. Moreover, \textit{TAG} implements progressive rendering or text annotations, making it possible to load in large documents without loss in rendering speed.

\textit{TAG} runs on any modern web browser and is available as open source software from \url{https://github.com/CreativeCodingLab/TextAnnotationGraphs}.

\section{Related Work}

\textit{TAG} is inspired by the \textit{BRAT} rapid annotation tool~\cite{stenetorp2012brat}. \textit{BRAT} is widely used for representing syntactic structure, but can also represent semantic events, and has been applied to a range of domain-specific NLP tasks, including biomedical data~\cite{verspoor2013annotating}. \textit{BRAT} supports a range of useful features that improve the overall efficiency of manual annotation tasks. However, \textit{BRAT} does not support the ability to draw links between links, which makes it difficult to represent relations linking several predicate-less relations, a feature necessary to completely describe complex events. Complex relations are previously explored by the authors in a range of different visualization projects that represent hierarchically-nested and/or clustered data derived from the machine reading of scientific texts describing biochemical events~\cite{biovisDang2,Dang2016a_IEEEEuroVis,Dang2017_Cactus_IEEEPacificVis,Dang2017_BioLinker_IEEEPacificVis,ForbesFontana_VAST2018,MurrayBMC2017,biovisPaduano111,Paduano2016_VINCI}.

\vspace{.1cm}

In addition to \textit{BRAT}, a range of newer projects investigate visual encodings for specific annotation tasks. For example, \newcite{Sarnat2017} introduce a web interface for exploring a parse tree. By entering in a sentence, an interactive visualization is created that includes expand/collapse functionality,
positional and color cues, explicit
visual support for sequential structure,
and dynamic highlighting to convey
node-to-text correspondence.  The tool includes an unusual representation of sequential structure as a container of linked nodes.  While this may help a user to distinguish relevant structural elements, it demands a large portion of the screen, which can make it difficult to view relationships over longer sequences of text. \textit{TAG} also utilizes a range of visual encodings to identify relationships and types, and includes an alternative representation of linguistic relationships. However, in \textit{TAG}, a user can fluidly switch between these representations or view them side by side.

\vspace{.1cm}


Another representation of a parse tree is used in the \textit{displaCy} software, introduced by~\newcite{displaCy}. \textit{displaCy}, however, is not equipped for annotation tasks and does not provide any interaction beyond the ability to scroll through a sentence (although more sophisticated functionality could potentially be built on top of it). By default, the entire body of text is placed on a single row, making it unwieldy for representing longer swaths of text.

\vspace{.1cm}

	\textit{WebAnno} is a flexible tool that supports multiple annotation
layers, and includes features to facilitate quality control, annotator management, and curation~\cite{WebAnno2,WebAnno3}. The visualrepresentation is similar to \textit{BRAT}, and the interface focuses mainly on resolving disagreeing annotations between users. \textit{WebAnno} includes a variety of built-in annotation layers, such as dependency relations, co-reference chains, and lemma forms, but in \textit{WebAnno} annotation expressiveness is limited in that it is not possible to create nested arcs that link to other arcs.

\vspace{.1cm}

\newcite{purgina2017visualizing} introduce \textit{WordBricks}, which also utilizes a container layout to identify linguistic structure while maintaining an explicit sequential representation that is easy to read. Although their approach eschews links that may lead to visual clutter, the \textit{WordBricks} layout can be difficult to read once the container is more than even a few levels deep, or in cases where the user wishes to quickly identify relationships between words or phrases across sentences or within longer passages.

\section{Text Annotation Graphs (TAG)}

In this section, we introduce the main visualization and interaction features of the \textit{TAG} software application. It consists of three panels: the annotation panel, the tree panel, and the options panel. Upon opening the software via any modern browser, \textit{TAG} shows the user a main annotation panel consisting of an example text snippet with both semantic and syntactic annotations. A secondary tree panel shows a ``semantic summary'' of a selected subgraph, in which text is organized using a network layout that emphasizes event relationships rather than text sequences. This alternative representation provides an overview of the event that can augment the more traditional reading of a parse tree. Additionally, it can be used to highlight the results of an event extraction process, and used as input for a structured search across other annotated documents. A third panel can be displayed on demand that provides the user with a range of options to change the visual appearance of text, to filter out particular annotation types, and to toggle on or off syntactic or semantic annotations.

\subsection{Loading Data}

	\textit{TAG} is meant to be agnostic to data schema and currently supports the following formats: BRAT standoff~\cite{stenetorp2012brat}, CoNLL-X~\cite{buchholz2006conll}, and bio-C~\cite{comeau2013bioc}. Additional formats can be added as desired by following the import templates for the supported formats. The software looks for files in a special `data' folder, and populates a drop down list to provide the option for the user to load in any of those data files. By choosing an item in this drop down list, which is located in the top left of the page, the data file is loaded into the \textit{TAG} application, and populates the annotation panel. A taxonomy file can also be associated with a data file, which enables \textit{TAG} to color annotations based on their taxonomic type, and makes it possible for users to search and filter by type.

\subsection{Annotation Panel}

The annotation panel takes up the majority of the \textit{TAG} application, filling up the entire page, or only the top two thirds of the page if the tree panel is opened. The annotation panel presents the text from the data file across many rows, requiring the user to scroll down to see text from longer passages. \textit{TAG} uses a progressive rendering strategy so that it is possible to display even large documents without loss in rendering speed. In each row, an annotation graph is shown both above and below the text. At the bottom, syntactic annotation is shown; At the top, semantic annotation is shown. Either syntactic or semantic annotations can be toggled on or off on demand.

   \begin{strip}
 
     \centering\noindent
     \includegraphics[width=6.8in]{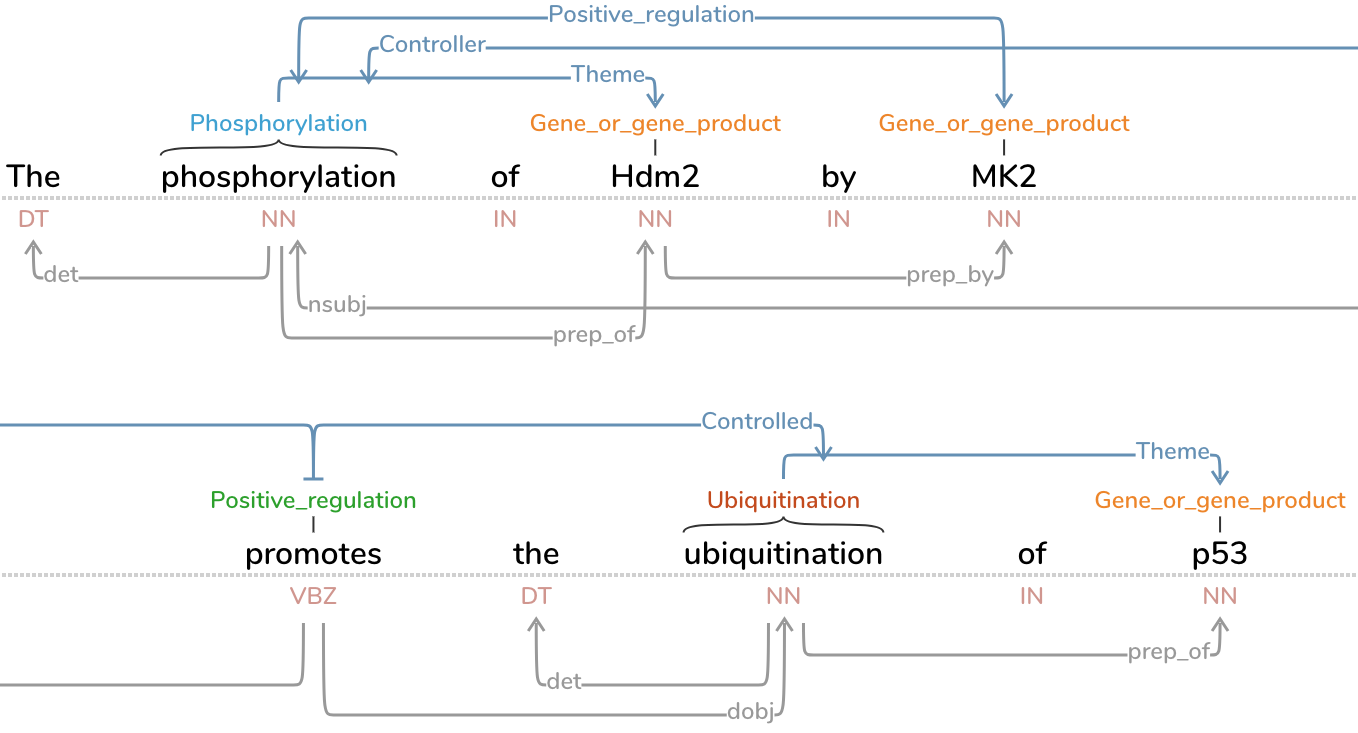} 
     \captionof{figure}{This screen capture shows the layout of a single sentence where annotation data is linked between elements which are presented on different rows. A user can interactively reposition words and links as desired across multiple rows in order to emphasize particular sections of text.}
     \label{tworows}
   \end{strip}

 \begin{strip}
 \vspace{1.0cm}
 
     \centering\noindent
     \includegraphics[width=6.8in]{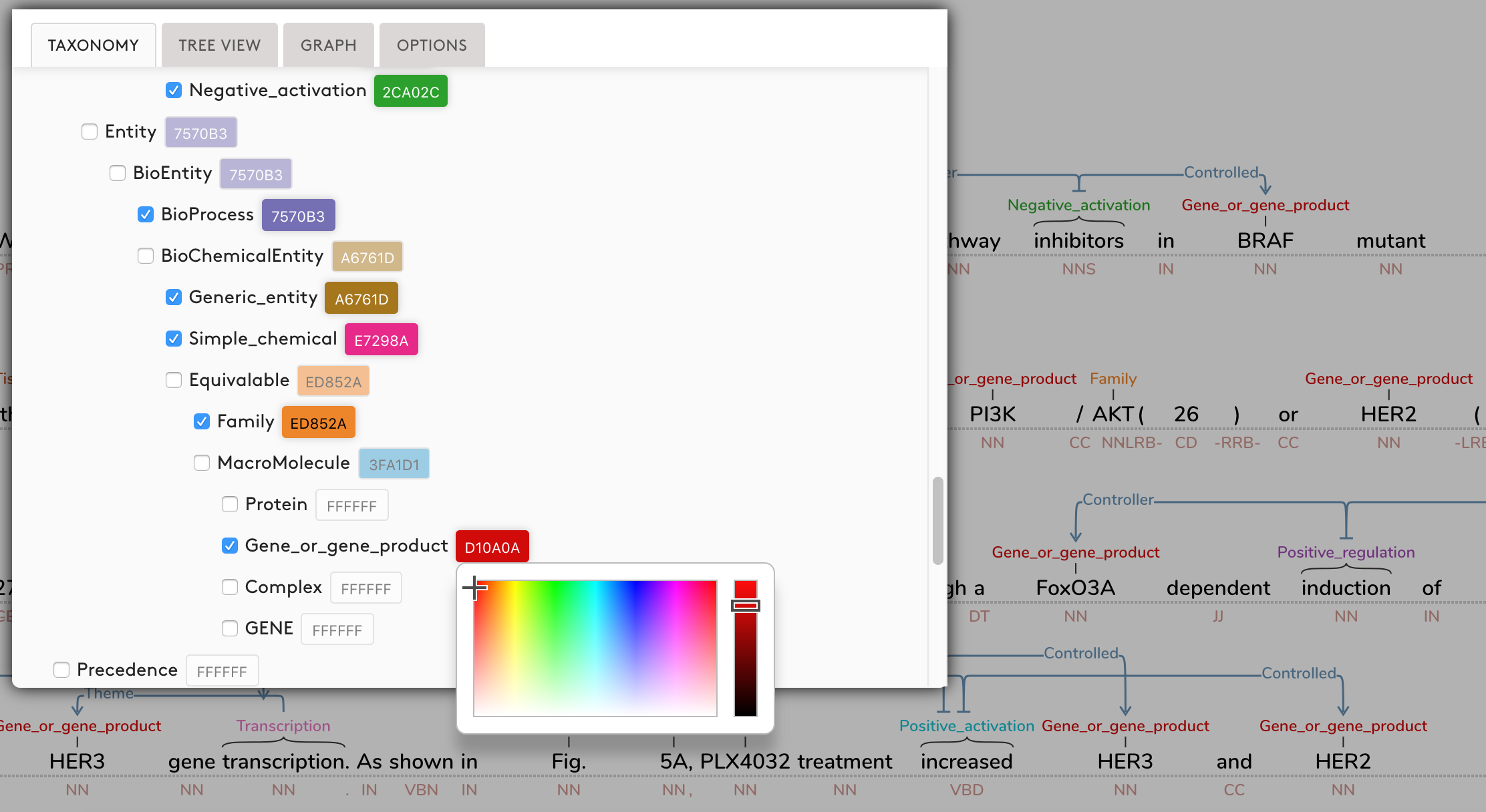} 
     \captionof{figure}{This screen capture shows a user updating the color of all nodes of type `Gene\_or\_gene\_product' via the interactive color picker. In the background, we can see that all examples of this annotation are updated dynamically. Selecting an annotation type that is a parent of `Gene\_or\_gene\_product', such as `MacroMolecule' or `Entity', would also update the color of `Gene\_or\_gene\_product', along with all its other children.}
     \label{taxonomyColors}
   \end{strip}

\textit{TAG} supports trigger-free relation mentions, providing the ability for the annotation graphs to utilize uni- or bi-directional links between nodes, links, or a combination of nodes and links. The layout of the links is organized using a custom algorithm that makes text labels on the graph easy to read, and minimizes edge crossings. Fig.~\ref{onerow} shows an example of data loaded into the annotation panel that requires only one row.  

\vspace{.1cm}

A user can interactively reposition the words in the text, including seamlessly moving words to different rows. This allows the user to emphasize particular word sequences by clustering a series of words closer together. Similarly, the vertices of links can be repositioned to improve the readability of the annotation graph. Fig.~\ref{tworows} show a view of an annotated sentence across two rows. Initially, it was positioned by default along a single row; here we see an example of how a user might choose to arrange the sentence interactively. In this case, annotation data is linked between elements which are on different rows and the syntax tree has been toggled on.  
\vspace{.1cm}

Annotations can be edited on demand. By clicking on a node, the user can select a type from a provided taxonomy, or freely type to replace the node's current annotation. Links can be repositioned so that one or more of the connections can instead connect to different link or node. A node or link can also be hidden or deleted, which also hides or removes any links directly connected to it.

\vspace{.1cm}

Additionally, a user can change the color of a single node (or all nodes of a particular type) by clicking on a node, which pops up an option panel, which allows the user to select a color dynamically. Figure~\ref{taxonomyColors} shows a user updating all nodes of type `Gene\_or\_gene\_product' to be colored red. When changing the color of a type, a user can also update all of its subtypes as well, if desired.

\begin{figure}[t!]
 \begin{center}
  \includegraphics[width=\columnwidth]{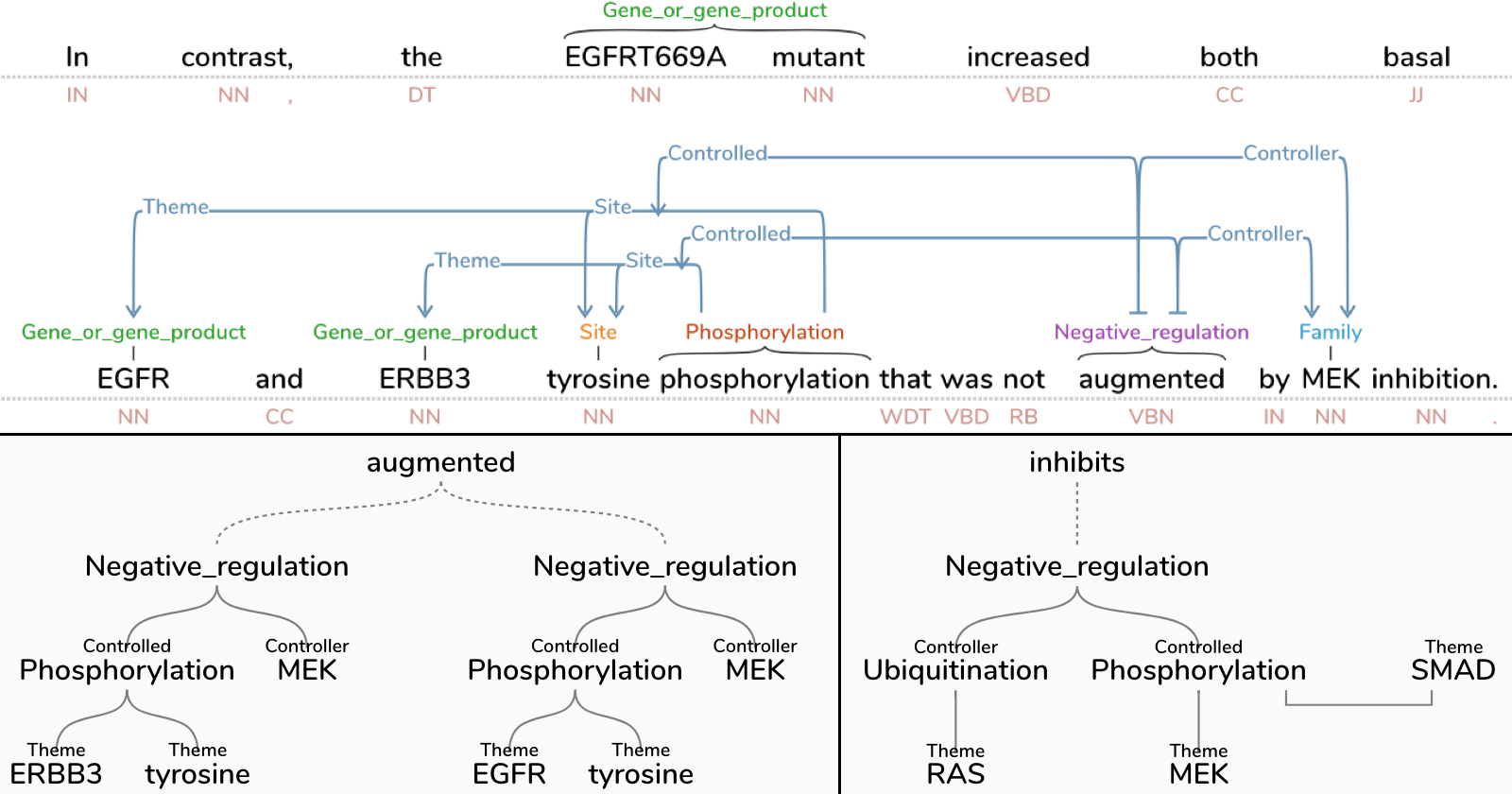} 
 \caption{The tree panel shows a semantic summary of a selected annotation. The top and bottom left show the same annotation in both formats, in the the tree panel emphasizes the negative regulation events triggered by the word `augmented.' The bottom right shows the semantic summary of the sentence shown in Fig.~\ref{onerow}, highlighting the events triggered by the word `inhibits.' }
 \label{treeFig}
 \end{center}
 \end{figure}

\subsection{Tree Panel}

By default the annotation panel fills up the entire screen, but the bottom third of the application can be used to provide an alternative representation of the annotated text. By double-clicking on any word or annotation, the tree panel appears in the application, displaying a tree whose root node is the element that was clicked on. The tree representation makes it easier to see the structure of the relations between annotation events, which may be more difficult to read in the annotation panel, especially when related events are spread across multiple sentences or rows, or when a single sequence of text contains multiple events. Of course, the tree representation removes the context from which the events are extracted. However, by presenting both views at once, the user is able to see the summary semantics of the annotation relations alongside the annotations embedded in the text.    

\vspace{.1cm}

Fig.~\ref{treeFig} shows an example of the tree panel. On the top, the annotation panel shows a sentence containing two negative regulation events whose representation has a somewhat dense layout, even after the user has repositioned the event so that all involved words are on the same row. On the bottom left, we see the simplified tree representation of the events, which appears when the user clicks on the word ``augments.'' On the bottom right, we see another tree representation--- the sentence described in Fig.~\ref{onerow}--- that highlights the annotations related to the word ``inhibits,'' and which shows an example of a sibling relationship.

\subsection{Option Panel}

The option panel provides various ways for the user to customize the software. The options include: the ability to toggle on or off the annotations above or below the text; optimization options, such as the ability to hide the annotation graph when repositioning words; and a mechanism for filtering annotations that contain (or do not contain) a specified taxonomic type. The user can also change the color of types defined in the taxonomy to highlight particular elements (a biomedical taxonomy can be seen in Fig.~\ref{taxonomyColors}).

\section{Examples}

Although \textit{TAG} can be used for a wide range of annotation tasks, here we highlight two examples that showcase situations that are difficult to represent using other software.

\subsection{Morphological Parse}

\begin{figure}[h!]
 \begin{center}
  \includegraphics[width=\columnwidth]{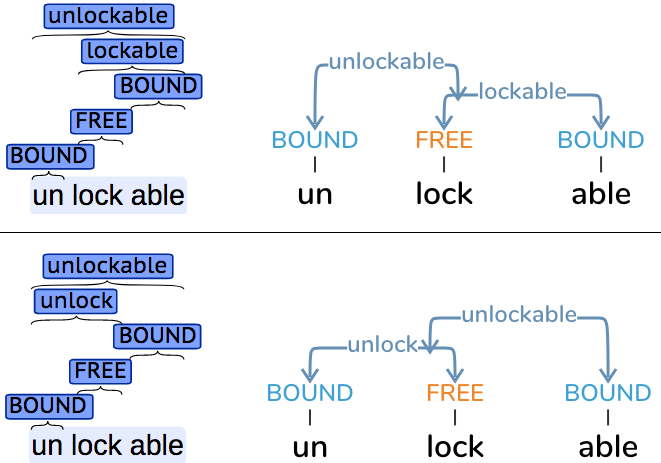} 
 \caption{Two interpretations for ``unlockable'' arising from differences in morphological derivation displayed in both \textit{BRAT} (left) and \textit{TAG} (right). The top parse models the interpretation ``cannot be locked'', while the bottom shows ``can be unlocked''. }
 \label{morphTrees}
 \end{center}
 \end{figure}

Fig.~\ref{morphTrees} shows a pair of examples that demonstrate semantic differences arising from morphological derivations of ``unlockable,'' which, as noted by~\newcite{larson1993interpreted}, could mean either ``cannot be locked'' (top) or ``can be unlocked'' (bottom). As we can see, trees such as these cannot easily be visualized in \textit{BRAT}, but can be represented in \textit{TAG}.

\subsection{Event Annotation}

\begin{figure}[t!]
 \begin{center}
  \includegraphics[width=\columnwidth]{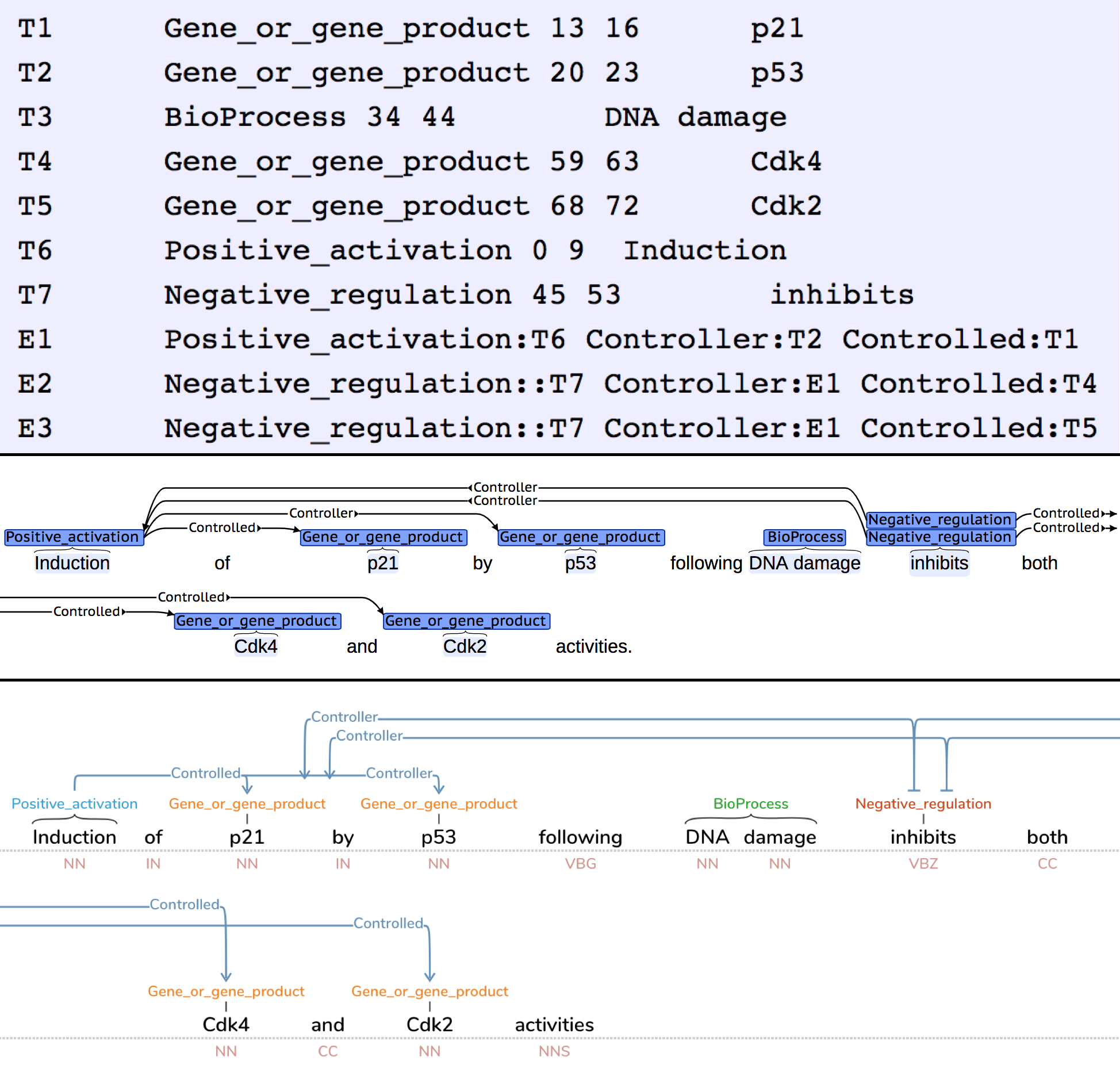} 
 \caption{Here we see screen captures of the same Reach events (top) in both \textit{BRAT} (middle) and \textit{TAG} (bottom). In \textit{TAG}, it is clear that E2 and E3 both use E1 as a controller of Cdk4 and Cdk2.}
 \label{brat_vs_tag_ea}
 \end{center}
 \end{figure}

Fig.~\ref{brat_vs_tag_ea} shows a comparison of how \textit{BRAT} and \textit{TAG} handle the output of Reach~\cite{Valenzuela2015domain} when processing the title text of the paper ``Induction of p21 by p53 following DNA damage inhibits both Cdk4 and Cdk2''~\cite{he2005}. Looking at extracted event E1, we can see that the Reach system finds that the positive activation event on the word ``Induction'' indicates p53 as the controller for p21. In the \textit{BRAT} representation, the ``controller'' and ``controlled'' relations are separated, whereas in \textit{TAG} it is easier to see that ``induction'' links p21 and p53 through a single positive activation trigger. Similarly, \textit{TAG} is able to recognize that the a single word serves as a trigger for two negative regulation events, and presents it as a single event with two simultaneous relations. Here, it is the entire positive activation event (E1) that serves as the controller for the inhibition of both Cdk4 and Cdk2.

\section{Conclusion}

This paper described the main functionality of the \textit{Text Annotation Graph} software, showcasing its increased expressivity for visually representing complex natural language phenomena for even large text documents. The software has been used so far to successfully represent events extracted from biomedical articles, but we expect that it could prove useful for annotation tasks across a range of domains. 

\vspace{.1cm}

Ongoing work is focused on extending \textit{TAG} to support comparative tasks, for instance, to use both the top and bottom half of each row to present the semantic annotations extracted from different systems. We are also investigating the use of \textit{TAG} to provide a human readable visualization of an active learning loop, enabled through saving ``diffs'' between the loaded data and the edited data. Since \textit{TAG} makes it easy to read and edit annotated events, we are exploring its use as a front end for crowdsourced user studies, such as those administered via Mechanical Turk, in order to gather data that can be used to train classifiers. 

\vspace{.1cm}

\textit{TAG} is written in JavaScript and runs in any browser. It makes extensive use of the SVG.js library,\footnote{\url{http://svgjs.com/}} which produces PDF quality output that is suitable for embedding in both online and printed documents. The software is open source and freely available from our GitHub code repository.\footnote{\url{https://github.com/CreativeCodingLab/TextAnnotationGraphs}}

\section*{Acknowledgements}

This work was funded by the DARPA Big Mechanism program under ARO contract W911NF-14-1-0395 and by the Bill and Melinda Gates Foundation HBGDki Initiative. 
Gus Hahn-Powell, Marco Valenzuela-Esc\'{a}rcega, and Mihai Surdeanu declare a financial interest in lum.ai. This interest has been properly disclosed to the University of Arizona Institutional Review Committee and is managed in accordance with its conflict of interest policies.


\section*{Bibliographical References}
\label{main:ref}

\bibliographystyle{lrec}
\bibliography{tags}


\end{document}